\documentclass[letterpaper, 10 pt, conference]{ieeeconf}  

\IEEEoverridecommandlockouts                              

\usepackage[utf8]{inputenc}
\usepackage[T1]{fontenc}
\usepackage{amsmath,amssymb,amsfonts}
\usepackage{graphicx}
\usepackage{textcomp}
\usepackage{xcolor}
\DeclareUnicodeCharacter{0308}{\"{}}
\usepackage{cite}
\usepackage{xurl}
\usepackage{subcaption}
\usepackage[table]{xcolor}   
\usepackage{colortbl}
\usepackage{booktabs}
\usepackage{multirow}
\usepackage{mathrsfs}
\usepackage[ruled,noend,linesnumbered]{algorithm2e}

\usepackage[colorlinks,linkcolor=black,anchorcolor=black,urlcolor=black,citecolor=black]{hyperref}

\usepackage[normalem]{ulem}
\useunder{\uline}{\ul}{}

\newcommand{\delete}[1]{{\bgroup\markoverwith{\textcolor{red}{\rule[0.5ex]{2pt}{0.4pt}}}\ULon{#1}}}

\title{
\LARGE \bf Primitive-based Truncated Diffusion for Efficient Trajectory\\Generation of Differential Drive Mobile Manipulators
}

\begin{document}
\author{Long Xu$^{\dagger}$, Choilam Wong$^{\dagger}$, Yuhang Zhong, Junxiao Lin, Jialiang Hou, and Fei Gao
\thanks{$^{\dag}$Indicates equal contribution.}
\thanks{E-mail: {\tt\small \{gaolon, fgaoaa\}@zju.edu.cn}}
}

\maketitle
\pagestyle{empty}

\begin{abstract}
We present a learning-enhanced motion planner for differential drive mobile manipulators to improve efficiency, success rate, and optimality. For task representation encoder, we propose a keypoint sequence extraction module that maps boundary states to 3D space via differentiable forward kinematics. Point clouds and keypoints are encoded separately and fused with attention, enabling effective integration of environment and boundary states information. We also propose a primitive-based truncated diffusion model that samples from a biased distribution. Compared with vanilla diffusion model, this framework improves the efficiency and diversity of the solution. Denoised paths are refined by trajectory optimization to ensure dynamic feasibility and task-specific optimality. In cluttered 3D simulations, our method achieves higher success rate, improved trajectory diversity, and competitive runtime compared to vanilla diffusion and classical baselines. The source code is released at \href{https://github.com/nmoma/nmoma}{\textcolor{magenta}{https://github.com/nmoma/nmoma}}.
\end{abstract}

\section{Introduction}
\label{sec:Introduction}
The development of embodied intelligence\cite{robocasa} has drawn significant attention to differential drive mobile manipulator (DDMoMa), which consists of multi-joint manipulator(s) and a differential drive base. This paper focuses on motion planning, a key component of autonomous navigation in DDMoMa that underpins the efficient and reliable execution of long-horizon tasks.


The objective of motion planning is to generate safe, dynamically feasible, and optimal trajectories guiding DDMoMa from an initial state to a target state. Prevailing methods\cite{zzd_icra,rampage,topay} rely on numerical optimization. Typical implementations obtain geometrically safe paths utilizing methods combining sampling and searching. The trajectory optimizer then initializes the optimization variables using this path to further derive a dynamically feasible trajectory that is optimal under specified criteria. 
In simple scenarios, path searching can be \delete{very fast} because the strategy can be greedy\cite{zzd_icra}. However, in complex scenarios, dense obstacles shrink the solution space, thus increasing the difficulty of path searching.

Recently, neural networks have been applied to path search problems due to their powerful representation capabilities and GPU-based parallel inference mechanisms, demonstrating remarkable success rates and efficiency on fixed-base manipulator(s)\cite{presto} and mobile platforms\cite{hzc_sr}. However, adapting them to DDMoMa remains challenging to achieve high real-time performance\cite{learn_neural,m2diffuser}. The primary challenges lie in designing effective and efficient network architectures and training frameworks. 

Compared with mobile bases or fixed-base manipulators, DDMoMa operates in a much larger 3D workspace and a substantially higher-dimensional configuration space. This requires the neural network to efficiently encode richer information. Moreover, the boundary states (i.e., the start and goal states of DDMoMa) and the environment constitute multimodal input, further complicating network design.

\begin{figure}[t]
    \centering
    \includegraphics[width=1.0\columnwidth]{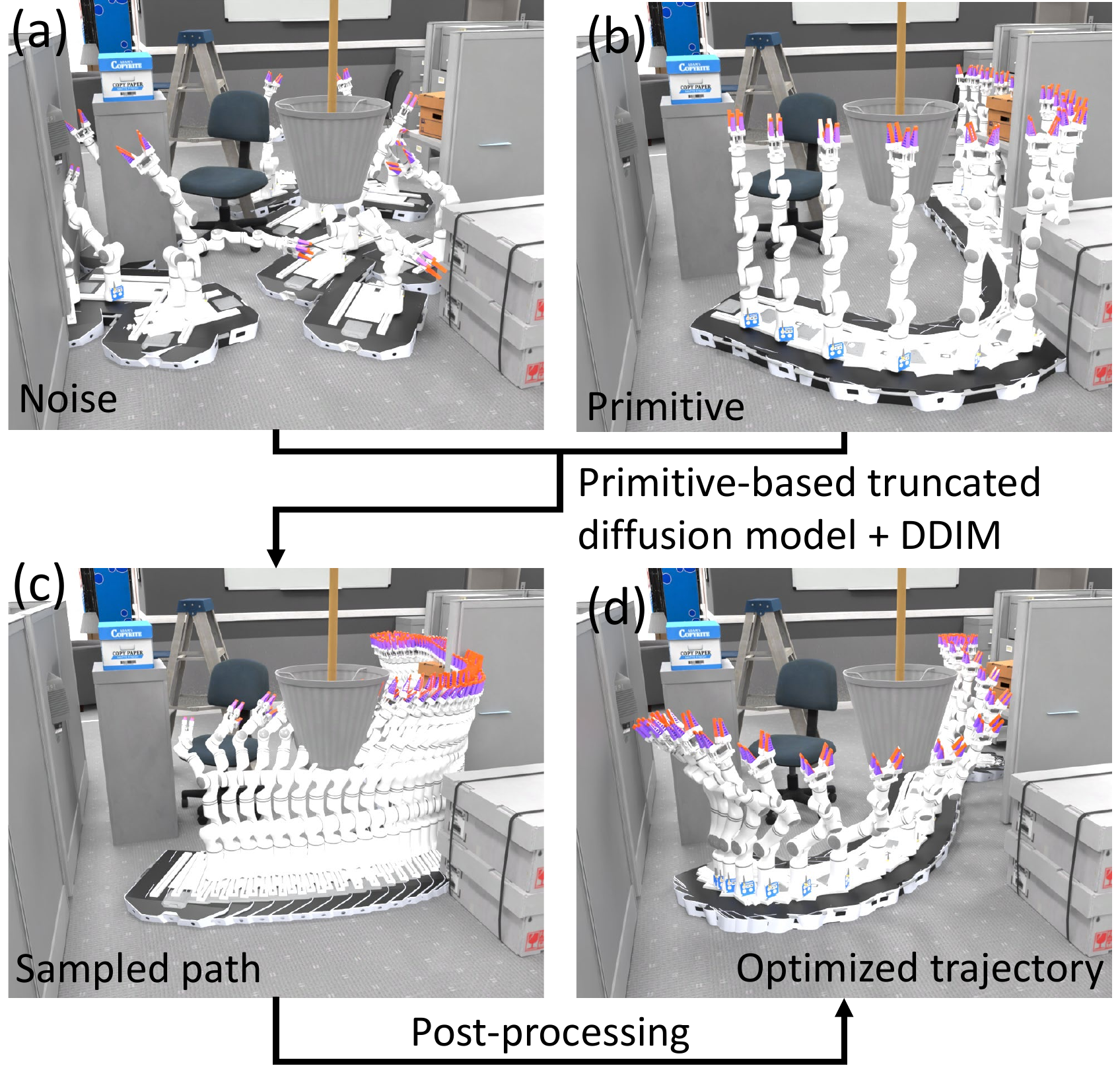}
	\caption{Process of the proposed planning framework during its deployment. Given the noise (subfigure (a)) and selected primitive (subfigure (b)), paths (subfigure (c)) are sampled via primitive-based truncated diffusion model (PTDM), subsequently post-processed to obtain the final optimized trajectory (subfigure (d)).}\label{fig:head}
    \vspace{-0.5cm}
\end{figure}

The DDMoMa path search problem is also characterized by numerous local optima. For example, when navigating around wall-like obstacles on the ground, identical manipulator motions may correspond to base paths with different topologies. In contrast, when avoiding suspended obstacles such as chandeliers, base paths with the same topology may still yield distinct manipulator trajectories, such as passing beneath or alongside the obstacle.
Diffusion models have recently demonstrated strong capabilities in capturing multi-modal distributions, making them a promising training framework for handling this problem. However, vanilla DDPM\cite{ddpm} has been shown to be susceptible to mode collapse\cite{diffusion_drive,mode_collapse}, leading to limited path diversity generated by the model. Such diversity is crucial in practice, as diverse initializations reduce the likelihood of the optimizer being trapped in poor local optima and thereby improve the probability of successful optimization\cite{topay}.

In this work, we construct a novel motion planning framework to address the aforementioned challenges. For network architecture,  to better fuse environment point clouds with the boundary states of the DDMoMa, we propose a keypoint sequence extraction (KSE) module embedded within the task representation encoder. This module maps the boundary states of DDMoMa into 3D space via differentiable forward kinematics. We then separately encode the point cloud and keypoint sequence, employing attention mechanisms for feature fusion. Compared to approaches that either omit KSE or extract features directly in 3D space, our method achieves more effective integration of point cloud and boundary states information. 

For training framework, we propose a primitive-based truncated diffusion model (PTDM) that generates DDMoMa paths from a biased distribution instead of sampling directly from a Gaussian distribution as done in the vanilla DDPM\cite{ddpm}. Simulation experiments demonstrate that the proposed strategy enhances both the efficiency and the diversity of generated paths. Fig.~\ref{fig:head} shows the data flow of the proposed framework at runtime. The contributions of the paper are:

\begin{itemize}
	\item A novel task representation encoder for DDMoMa motion planning. It combines attention mechanisms with the proposed differentiable keypoint sequence extraction, enabling effective feature fusion for environment point clouds and the boundary states of DDMoMa.
	\item A primitive-based truncated diffusion model for DDMoMa motion planning, enhancing the efficiency and path diversity.
	\item Integrating the above two modules with a optimization-based post-processing, we propose a learning-enhanced planning framework for DDMoMa. Comprehensive evaluations in three types of cluttered 3D simulation environments validate the efficiency of the pipeline.
\end{itemize}

\section{Related Works}
\subsection{Classical Motion Planning}
Classical motion planning pipelines typically consist of frontend and backend. The frontend is responsible for generating a collision-free path, which serves as an initial guess for the backend. The backend applies trajectory optimization to refine this path, incorporating higher-order dynamics to produce a trajectory that is dynamically feasible. 

Sampling-based methods\cite{ompl} are typically used as the frontend for motion planning of the DDMoMa. They probe the state space with random sampling and incrementally construct a graph representation approximating the free state space, thereby converting path search into a graph search problem. Although they offer desirable theoretical guarantees, such as probabilistic completeness and asymptotic optimality, they are susceptible to the curse of dimensionality\cite{mpnet}. The volume of the configuration space to be approximated grows exponentially with the degrees of freedom (DoF), causing sampling time to become prohibitively long for high-DoF systems, such as DDMoMa, resulting in the difficulty of real-time planning.

To improve the efficiency of the frontend, recent methods\cite{rampage,zzd_icra,topay,momatase} often decompose and plan the paths of mobile base and the mounted robotic arm separately. Yang et al.\cite{rampage} employ a topo-guided kinodynamic searching method to find the base path. Based on it, they propose a spatiotemporal rapidly exploring random tree to further obtain the whole-body path. Similarly, Wu et al.\cite{zzd_icra} propose a multilayer constrained RRT*-Connect method to sample the manipulator state for each waypoint from the base path searched by Hybrid-A*. To further improve the efficiency and success rate, Xu et al.\cite{topay} adopt the concept of uniform visibility deformation class\cite{topo} to generate topological paths, parallel sampling the paths of the manipulator for different base paths. They also propose a novel polynomial-based trajectory representation for efficient trajectory optimization.

\subsection{Learning-enhanced Motion Planning}
While the aforementioned greedy decomposition strategy enables real-time motion planning for DDMoMa, it still relies on full state-space sampling as a fallback. In highly complex environments, it inevitably resorts to this time-consuming process\cite{topay}.


Neural networks’ powerful representation capability and natural suitability for parallel computation enable efficient motion planning in complex scenarios.
A prominent strategy for accelerating sampling-based planners is to bias the sampler with task and environment-specific prior knowledge\cite{mpnet,ichter2018learning,learn_neural}. Since the space of feasible paths is highly constrained, concentrating sampling efforts on promising regions yields considerable efficiency gains over a uniform sampling approach.
For instance, Ichter et al. \cite{ichter2018learning} utilizes conditional variational autoencoder (CVAE) to learn and sample from the distribution of waypoints conditioned on the planning task and environment.

Similarly, Neural Randomized Planner \cite{learn_neural} enhances classical sampling-based methods with a learning-based sampler, aiming to prioritize sampling in promising region given the local environment. 
Aside from improved efficiency, another key advantage of this enhanced state space sampling paradigm is that, both probabilistic completeness and asymptotic optimality can be retained by combining learned sampler and uniform sampler. However, striking a balance in sampling quality and efficiency is challenging. While an expressive enough model that yields high-quality waypoints could be time consuming, a fast and light-weight model could render subsequent graph search difficult. In addition, adopting the state space paradigm necessitates computationally expensive graph construction and graph search.




\begin{figure*}[t]  
		\centering
		{\includegraphics[width=1.0\linewidth]{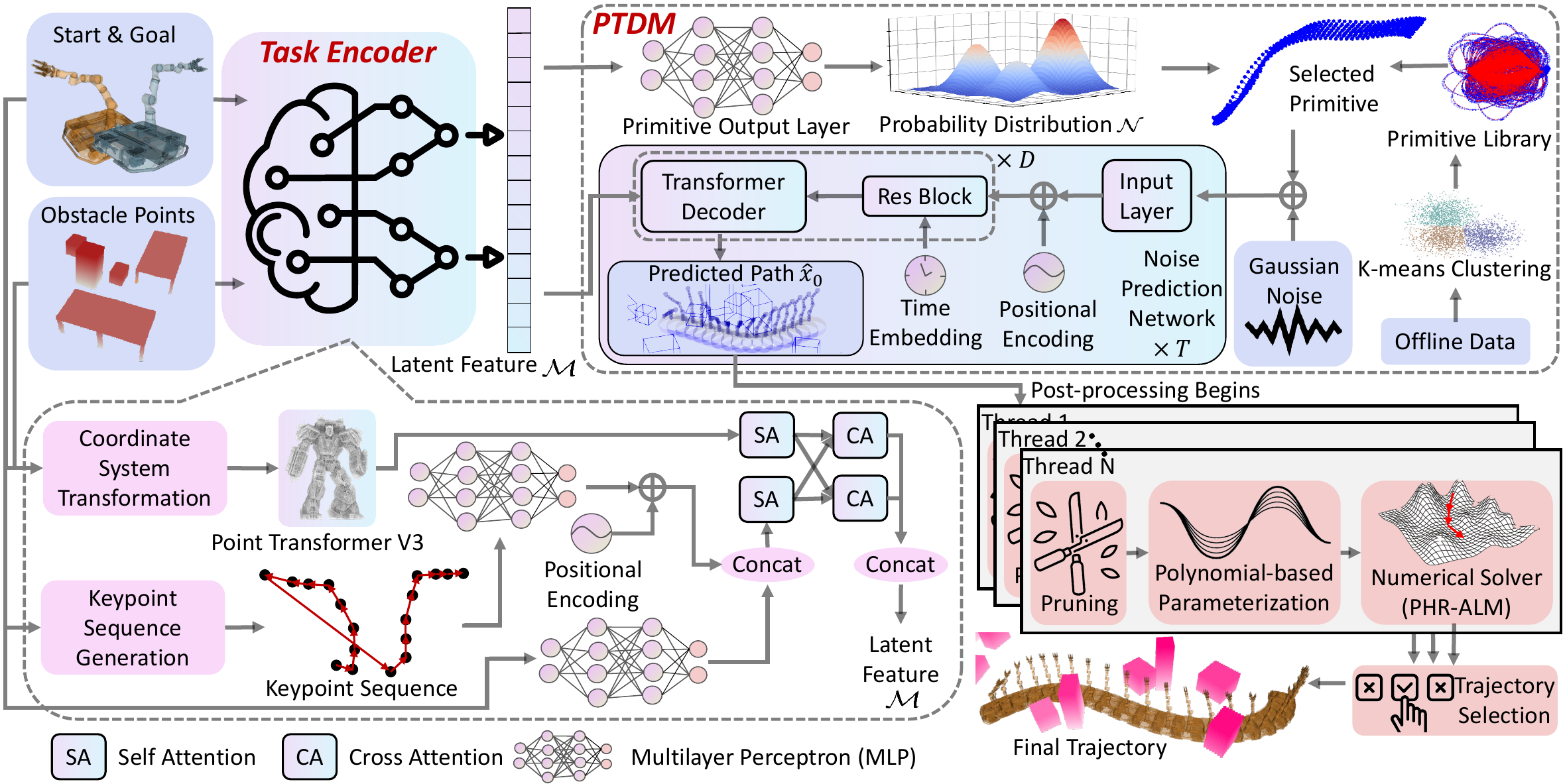}}
		\caption{Proposed planning framework. The neural network encodes the task and samples robot paths efficiently. The paths are then refined by a model-based trajectory optimizer to generate safe, dynamically feasible trajectories.}
		\label{fig:framework}
\end{figure*}

In recent years, diffusion models have emerged as a powerful tool for motion planning, capitalizing on their proven ability to capture multi-modal distributions in high-dimensional spaces\cite{dp,cdm,m2diffuser,MPD2025,priordm,diffusion_drive}. For instance, M$^2$ Diffuser\cite{m2diffuser} learns distribution of trajectory conditioned on the environment, and applies guided sampling to sample from the posterior distribution. Similarly, MPD\cite{MPD2025} learns the marginal distribution and samples from the posterior distribution by enforcing physical and task-related constraints through guided sampling. Such approaches typically require the injection of guidance signals in multiple denoising steps, making it difficult to leverage acceleration techniques such as DDIM\cite{ddim}. The need for tens or even hundreds of denoising iterations severely limits their real-time applicability. For example, M$^2$ Diffuser\cite{m2diffuser} requires over $50$ guided denoising steps, resulting in planning times of several seconds per query. To obviate the need for guided sampling and thereby enable the use of acceleration techniques for faster inference, Seo et al.\cite{presto} introduce trajectory optimization as post-processing. Their diffusion model learns directly from the posterior distribution, demonstrating efficient motion planning of a fixed-base manipulator in a shelf scenario.

Although acceleration techniques such as DDIM\cite{ddim} can be employed to speed up the inference of vanilla diffusion models, a reduced number of sampling steps often results in performance degradation. To mitigate this issue, Liao et al.\cite{diffusion_drive} propose an anchor-based truncated diffusion approach, which learns the denoising process from the anchored Gaussian distribution. They demonstrate a balanced trade-off between efficiency and effectiveness in trajectory generation for autonomous driving. Our method is also inspired by the truncated diffusion model\cite{truncated_diffusion}. However, unlike~\cite{diffusion_drive}, our method determines the primitive independently of the diffusion process and prior to its execution. We construct a distribution based on the primitive for the denoising process, rather than denoising the anchored Gaussian distribution composed of all primitives. This approach provides the denoising network with a clearer initial distribution and eliminates the need to sample trajectories for all primitives, thereby reducing computational cost.

\section{Planning Framework}
Fig.~\ref{fig:framework} illustrates our planning framework. Upon receiving the obstacle points $\boldsymbol{P}_e\in\mathbb{R}^{N_{e}\times3}$ and boundary states $\boldsymbol{s}_s,\boldsymbol{s}_g\in SE(2)\times\mathbb{R}^{N_{m}}$, the robot analyses the task with the proposed task encoder, where $N_e$ denotes the number of points, $N_{m}$ denotes the DoF of the manipulator. After selecting motion primitive to sample paths based on latent feature $\mathcal{M}$, it performs model-based post-processing to obtain dynamically feasible trajectories.

\subsection{Task Representation Encoder}
The task encoder consists of three components: differentiable preprocessing, feature encoding, and feature fusion, as shown in the bottom left part of the Fig.~\ref{fig:framework}. We perform a coordinate system transformation on the point cloud. Let the start position of the mobile base be $\boldsymbol{p}_s=[x_s,y_s]$ and the goal position be $\boldsymbol{p}_g=[x_g,y_g]$. We define the rotation matrix $\boldsymbol{R}_e\in\mathbb{R}^{3\times3}$ as follows:
\begin{align}
    \theta_d&=\text{atan2}(y_g-y_s,x_g-x_s),\\
    \boldsymbol{R}_e&=\left[
    \begin{array}{ccc}
    \cos\theta_d& -\sin\theta_d&0\\
    \sin\theta_d& \cos\theta_d&0\\
    0&0&1
    \end{array}\right].
\end{align}
Thus, the transformed point cloud is $\boldsymbol{P}_t=(\boldsymbol{P}_s-[\boldsymbol{p}_s,0])\boldsymbol{R}_e$. 

\begin{figure}[t]  
		\centering
		{\includegraphics[width=1.0\columnwidth]{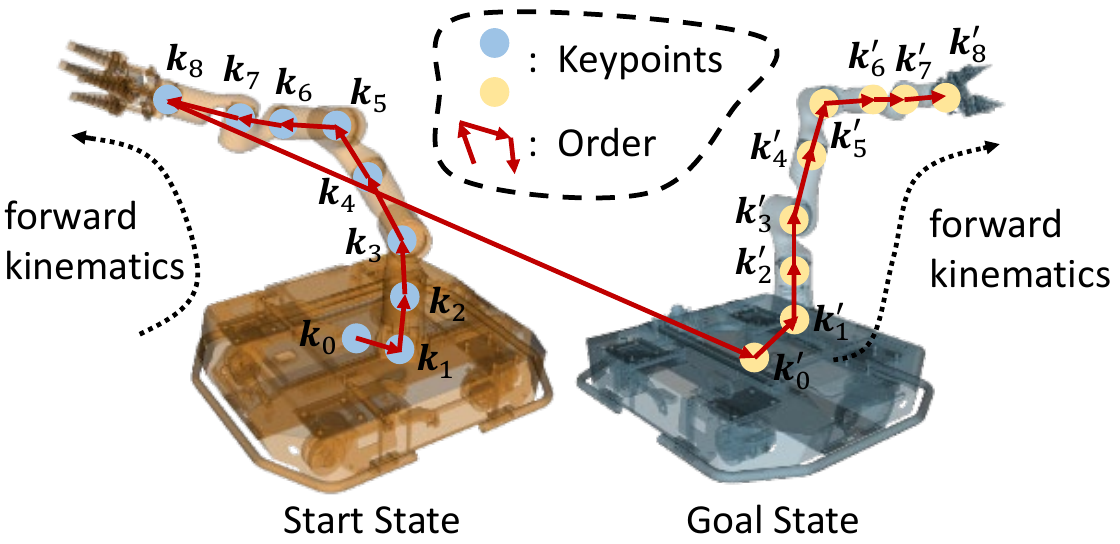}}
		\caption{An example of key point sequence generation for DDMoMa consisting of a 7-DoF manipulator and a mobile base. }
		\label{fig:kse}
\end{figure}

For robot boundary states, using the proposed keypoint sequence extraction (KSE) module, we map them into the 3D space where the point cloud lies, yielding a sequence of keypoints. As shown in Fig.\ref{fig:kse}, the keypoints correspond to the joint coordinates, the center of the mobile base, and the fixed center of the manipulator. Let the state of the robot $\boldsymbol{s}=[x,y,\theta,\boldsymbol{q}^\text{T}]^\text{T}\in SE(2)\times\mathbb{R}^{N_{m}}$, where $\boldsymbol{q}=[q_1,q_2,...,q_{N_{m}}]^\text{T}\in\mathbb{R}^{N_{m}}$ represent the joint angles of the manipulator. We obtain keypoints via differentiable forward kinematics with the following equations:
\begin{align}
    \boldsymbol{k}_0&=[x,y,0,\cos\theta]\triangleq[\boldsymbol{t}_0,c_\theta],\\
    \boldsymbol{k}_1&=[\boldsymbol{l}_1\boldsymbol{R}_\theta+\boldsymbol{t}_0,\sin\theta]\triangleq[\boldsymbol{t}_1,s_\theta],\\
    \boldsymbol{k}_2&=[\boldsymbol{l}_2+\boldsymbol{t}_1,q_{n1}]\triangleq[\boldsymbol{t}_2,q_{n1}],\\
    \boldsymbol{k}_3&=[\boldsymbol{l}_3\boldsymbol{R}_{q_1}+\boldsymbol{t}_2,q_{n2}]\triangleq[\boldsymbol{t}_3,q_{n2}],\\
    &\vdots\nonumber\\
    \boldsymbol{k}_j&=[\boldsymbol{l}_j\boldsymbol{R}_{q_{j-2}}+\boldsymbol{t}_{j-1},q_{n(j-1)}]\triangleq[\boldsymbol{t}_j,q_{n(j-1)}],\\
    &\vdots\nonumber\\
    \boldsymbol{k}_{N_{m}+1}&=[\boldsymbol{l}_{N_{m}+1}\boldsymbol{R}_{q_{N_{m}-1}}+\boldsymbol{t}_{N_{m}},q_{nN_{m}}],
\end{align}
where the first three dimensions of each keypoint $\boldsymbol{k}_i,i=0,1,...,N_{m}+1$ represent the coordinates, while the last dimension denotes the feature. $\boldsymbol{R}_*,*=\{\theta,q_1,...,q_{N_{m}-1}\}$ denote rotation matrices corresponding to rotations about the Z-axis or Y-axis of the keypoint coordinate system. $\boldsymbol{l}_1\sim\boldsymbol{l}_{N_{m}+1}$ are link vectors. Referencing the continuous representation\cite{rotation6d} for $SE(2)$, we set their features as $\cos\theta$ and $\sin\theta$ respectively. $q_{ni},i=1,2,...,N_{m}$ are also normalized to $[-1,1]$ as follows.
\begin{align}
       q_{ni}=\frac{2q_i-q_{i\text{max}}-q_{i\text{min}}}{q_{i\text{max}}-q_{i\text{min}}},
\end{align}
where $q_{i\text{min}},q_{i\text{max}}$ are the joint limitations. The order of the keypoint sequence is defined from the base to the end effector, from the start state to the goal state.

Although the keypoints are mapped into the same 3D space as obstacle points, their physical meaning remains fundamentally different. Thus, we employ different encoders to extract their features. Multilayer Perceptron (MLP) and Point Transformer V3\cite{ptrans3} are adopted as keypoint encoder and point encoder, respectively. We utilize self-attention and cross-attention mechanisms for feature fusion, where sinusoidal positional encoding is applied to the keypoint sequences for introducing order information. To enrich the information prior to feature fusion, we also apply MLP directly to robot boundary states, concatenating the results with the keypoint embeddings.

\subsection{PTDM for DDMoMa}
Truncated diffusion model has been demonstrated to be a faster and computationally cheaper approach than classical diffusion model\cite{ddpm,ddim} in the fields of image generation\cite{truncated_diffusion} and autonomous driving\cite{diffusion_drive}. Drawing inspiration from them, in order to improve efficiency and prevent mode collapse\cite{mode_collapse}, we propose a primitive-based truncated diffusion model (PTDM) for motion planning of DDMoMa. 

In our work, the path of DDMoMa is expressed as a sequence of states $\boldsymbol{\tau}=[\boldsymbol{\mathfrak{s}}_1,\boldsymbol{\mathfrak{s}}_2,...,\boldsymbol{\mathfrak{s}}_{N_{\boldsymbol{\tau}}}]\in\mathbb{R}^{(4+N_{m})\times N_{\boldsymbol{\tau}}}$, where $\boldsymbol{\mathfrak{s}}_i=[x_i,y_i,c_{\theta i},s_{\theta i},\boldsymbol{q}_i^\text{T}]^\text{T}\in\mathbb{R}^{4+N_m},i=1,2,...,N_{\boldsymbol{\tau}}$.$N_{\boldsymbol{\tau}}$ is the number of states. As shown in the top right part of the Fig.~\ref{fig:framework}, the PTDM generates the samples by learning the denoising process $p_\psi(\boldsymbol{\tau}^{t-1}|\boldsymbol{\tau}^t,\boldsymbol{c})$ from a noisy-primitive distribution to the original data distribution $p(\boldsymbol{\tau}^0|\boldsymbol{c})$, where the conditions $\boldsymbol{c}$ include $\boldsymbol{P}_e,\boldsymbol{s}_s,\boldsymbol{s}_g$. 

The noise distribution predicted by the neural network can be expressed as:
\begin{align}
    p_\psi(\boldsymbol{\tau}^0|\boldsymbol{c})=\int p(\boldsymbol{\tau}^{\tilde{T}}|\boldsymbol{c})\prod_{t=1}^{\tilde{T}}p_\psi(\boldsymbol{\tau}^{t-1}|\boldsymbol{\tau}^t,\boldsymbol{c})d\boldsymbol{\tau}^{1:\tilde{T}},
\end{align}
where $p(\boldsymbol{\tau}^{\tilde{T}})$ is the noisy-primitive distribution. We generate it by truncating the diffusion noise schedule and fusing the truth primitive with Gaussian distribution:
\begin{align}
    p(\boldsymbol{\tau}^{\tilde{T}})&=\sqrt{\bar\alpha^{\tilde{T}}}\boldsymbol{\tau}_\text{prim}+\sqrt{1-\bar{\alpha}^{\tilde{T}}}\boldsymbol{\epsilon},\quad\boldsymbol{\epsilon}\sim\mathcal{N}(0,\boldsymbol{I}),\label{eq:truncT}
\end{align}
where $\bar{\alpha}^t=\prod_{i=1}^t\alpha^i,\alpha^i\in(0,1),i\in[1,\tilde{T}]$ are predefined parameters for the noise schedule. $\tilde{T}<T_\text{max}$ is the number of truncated diffusion steps, where $T_\text{max}$ is the number of original diffusion steps. $\boldsymbol{\tau}_\text{prim}$ is the truth primitive, the one in the primitive library with the closest Euclidean distance to the truth path $\boldsymbol{\tau}^0$. Here, we employ the K-Means clustering algorithm for primitive library construction from offline-generated trajectories, which have been demonstrated to be practically effective for primitive generation.\cite{diffusion_drive,e2e_fly}

During training, the primitive output layer predicts the probability distribution of the primitives. The target distribution for noise prediction network is
\begin{align}
    q(\boldsymbol{\tau}^{t}|\boldsymbol{\tau}^{t-1},\boldsymbol{c})=\mathcal{N}(\sqrt{\alpha^t}\boldsymbol{\tau}^{t-1},(1-\alpha^t)\boldsymbol{I}).
\end{align}

\subsection{Post-processing}


To generate dynamically feasible trajectories for DDMoMa, we adopt an efficient trajectory representation based on piecewise polynomials and arc length-yaw parameterization\cite{topay}. Initializing the trajectories with paths sampled using PTDM, we perform trajectory optimization in parallel, as shown in the bottom right part of Fig.~\ref{fig:framework}. 

\begin{figure*}[t]  
		\centering
		{\includegraphics[width=1.0\linewidth]{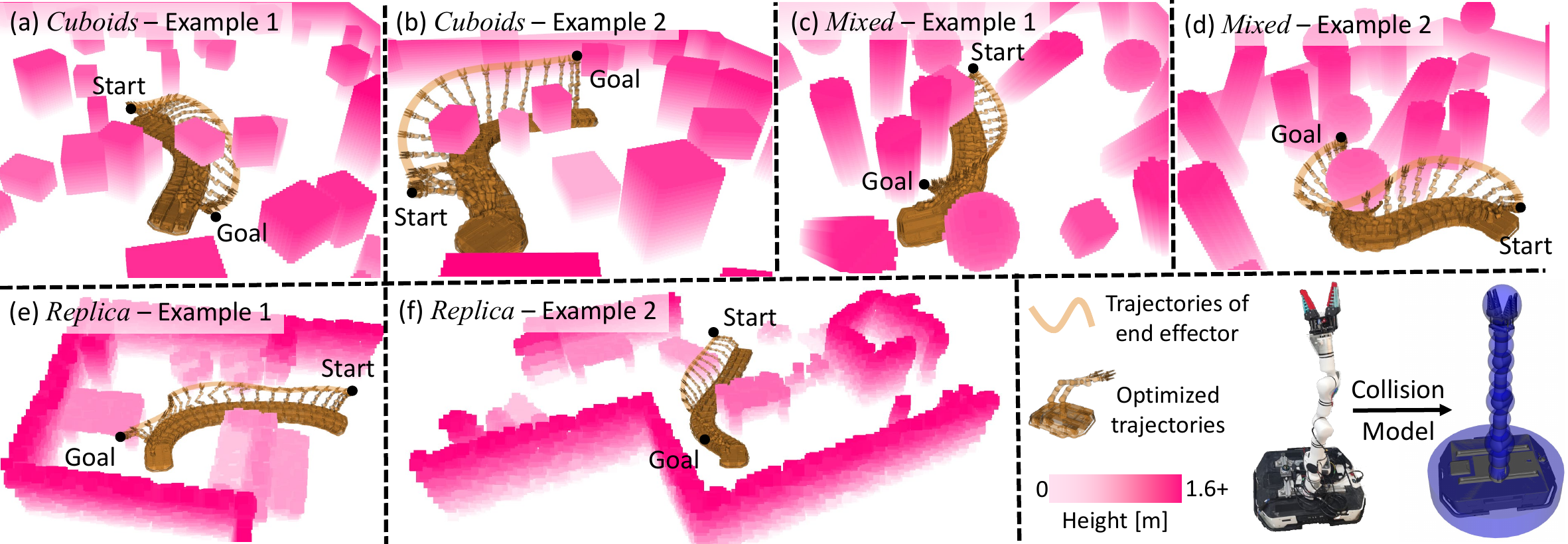}}
		\caption{Robot configuration and simulation environments. Two examples are provided for each environment here, with trajectories generated by the proposed planning framework.}
		\label{fig:sim}
\end{figure*}

The trajectory optimization formulation is identical to that in the work~\cite{topay}, except that the optimization variables for goal state of the mobile base is changed to $[a_f,\theta_f]$, and the constraint is modified to
\begin{align}
    x_g&=x_s+\int_0^{t_{f}}\dot{a}(\tau)\cos\theta(\tau)d\tau,\\
    y_g&=y_s+\int_0^{t_{f}}\dot{a}(\tau)\sin\theta(\tau)d\tau,
\end{align}
where $a(t),\theta(t)$ are the arc length and yaw trajectories, respectively. $t_f>0$ is the total duration of the trajectory. $a_f=a(t_f),\theta_f=\theta(t_f)$.

To obtain better initial values, inspired by the pruning algorithm in OMPL\cite{ompl}, we perform a pruning step on the sampled paths before optimization. 
We also adopt the early termination strategy proposed in TopAY\cite{topay} to improve efficiency.

\section{Implementation Details}
\label{sec:implement}
\subsection{Training Losses}
To facilitate explicit geometric losses, we choose to directly predict the original path $\boldsymbol{\tau}^0$, instead of the noise $\boldsymbol{\epsilon}$. In this work, the added geometric loss is defined as the expected loss form of the predicted path $\hat{\boldsymbol{\tau}}$: $\bar{\mathcal{L}}_{geo}(\hat{\boldsymbol{\tau}})=\mathbb{E}_{\boldsymbol{\tau}_\text{prim}\sim P_{\psi}(\boldsymbol{\tau}_\text{prim})}[\mathcal{L}_{geo}(\hat{\boldsymbol{\tau}})]=\sum_{\boldsymbol{\tau}_\text{prim}}\mathcal{L}_{geo}(\hat{\boldsymbol{\tau}})P_{\psi}(\boldsymbol{\tau}_\text{prim})$, where $\mathcal{L}_{geo}(\hat{\boldsymbol{\tau}})$ is the weighted sum of following losses.
\begin{align}
    \mathcal{L}_{safe}(\hat{\boldsymbol{\tau}})&=\sum_{i=1}^{N_{\boldsymbol{\tau}}}\sum_{j=1}^{N_{c}}\frac{\text{ReLU}(r_j^{thr}-\text{SDF}(\textbf{ColliPts}(\hat{\boldsymbol{\mathfrak{s}}}_i,j)))}{N_{\boldsymbol{\tau}}},\nonumber\\
    \mathcal{L}_{smooth}(\hat{\boldsymbol{\tau}})&=\sum_{j\in\mathcal{S}}\text{ReLU}(\sum_{i=1}^{N_{\boldsymbol{\tau}}-1}\Delta_{j,i}-\sum_{i=1}^{N_{\boldsymbol{\tau}}-1}\Delta_{j,i}^*),\label{eq:smooth}\\
    \mathcal{L}_{unip}(\hat{\boldsymbol{\tau}})&=\frac{\text{Std}([\Delta_{p,1},\Delta_{p,2},...,\Delta_{p,N_{\boldsymbol{\tau}}}])}{\sqrt{N_{\boldsymbol{\tau}-1}\sum_{i=1}^{N_{\boldsymbol{\tau}}-1}\Delta_{p,i}^*}},\label{eq:unip}
\end{align}
where $\mathcal{L}_{safe}$ is obstacle avoidance loss. The function $\textbf{ColliPts}(\boldsymbol{x},y)$ maps the state $\boldsymbol{x}\in\mathbb{R}^{4+N_m}$ and index $y\in\mathbb{Z}_+$ to collision detection point $\boldsymbol{c}_{\boldsymbol{x},y}\in\mathbb{R}^3$. As shown by the robot collision model in Fig.~\ref{fig:sim}, we approximate the collision geometry using a set of cylinders and spheres. $r_j^{thr}$ is the safe threshold corresponding to the $j$-th point. The function $\text{SDF}:\mathbb{R}^3\mapsto\mathbb{R}$ is Signed Distance Field (SDF) corresponding to the environment. In SDF, the value at each state of the space is the distance to the edge of the nearest obstacle, which is negative inside the obstacles.

Eq.(\ref{eq:smooth}) denotes smoothness loss, where $\mathcal{S}=\{p,\theta,\boldsymbol{q}\}$. We represent smoothness by combining the arc length of the base path, the total change in the yaw angle of the base, and the joint angles of the manipulator.
\begin{align}
    \Delta_{p,i}&=\sqrt{(x_{i+1}-x_i)^2+(y_{i+1}-y_i)^2},\\
    \Delta_{\theta,i}&=\sqrt{(c_{\theta(i+1)}-c_{\theta i})^2+(s_{\theta(i+1)}-s_{\theta i})^2},\\
    \Delta_{\boldsymbol{q},i}&=\sum_{j=1}^{N_m}\vert q_{j,i+1}-q_{j,i}\rvert.
\end{align}
$\Delta_{j,i},\Delta_{j,i}^*$ denote the values that correspond to the predicted path $\hat{\boldsymbol{\tau}}$ and the gourd truth $\boldsymbol{\tau}^0$.

Eq.(\ref{eq:unip}) denotes the uniform loss. In this work, we perform equal-interval arc length sampling on the trajectory optimized by classical trajectory planner to generate the ground truth. Thus, we penalize the normalized standard deviation to meet this condition. In addition, focal loss\cite{focal_loss} is chosen for primitive classification. 

\subsection{Data Collection}
Following the work~\cite{topay}, we generate hundreds of random environments and random start-goal states to construct motion planning tasks. For data collection, we employ an efficient planner TopAY\cite{topay} and adjust its parameters to allow sufficient time for path searching and trajectory optimization. To compensate for incompleteness of the front end in TopAY\cite{topay}, we adopt OMPL\cite{ompl} to sample the path in full state space of DDMoMA when TopAY\cite{topay} fails. Optimized trajectories are checked to ensure dynamic feasibility and safety, then recorded with corresponding boundary states, obstacle point clouds, and SDF.

\begin{figure*}[t]  
		\centering
		{\includegraphics[width=1.0\linewidth]{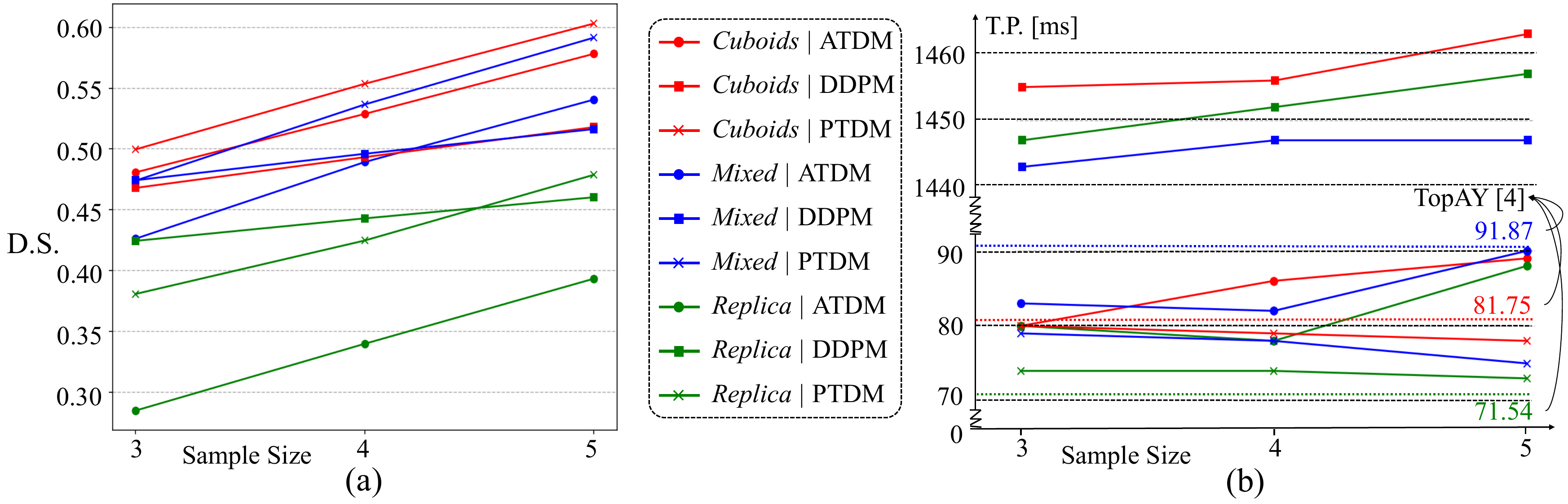}}
		\caption{Comparisons of diversity score (D.S., subfigure (a)) and planning time (T.P., subfigure (b)) across different diffusion strategies as the number of sampled paths varies in different environments. In most of cases, PTDM achieves lower planning time consuming and higher diversity than vanilla DDPM and anchor-based diffusion (ATDM). The \textcolor[rgb]{1.0,0.0,0.0}{red}, \textcolor[rgb]{0.0,0.0,1.0}{blue}, and \textcolor[rgb]{0.0,0.5,0.0}{green} dashed lines in subfigure (b) represent the average T.P. of TopAY\cite{topay} in \textit{Cuboids}, \textit{Mixed}, and \textit{Replica}, respectively.}
		\label{fig:line_ptdm}
\end{figure*}

\section{Results}
\subsection{Experimental Setup}
The DDMoMa used in TopAY\cite{topay} is employed in this work to evaluate the proposed method. It consists of a two-wheeled differential drive mobile base and a 7-DoF manipulator, as shown in the bottom right corner of  Fig.~\ref{fig:sim}. We employ three different kinds of scenes in simulation for benchmark comparisons, which are denoted by \textit{Cuboids}, \textit{Mixed}, and \textit{Replica}, as shown in Fig.~\ref{fig:sim}(a)-(f). Each are $10m\times10m$ walled rooms filled with randomly placed obstacles of random size. For \textit{Cuboids}, there are $20$ grounded cuboids and $30$ floating cuboids. To increase the environmental complexity, we introduce additional types of geometric obstacles, forming the \textit{Mixed}, which includes $10$ cuboids, $10$ spheres, and $25$ cylinders. To perform testing in more realistic scenarios, we adopt the ReplicaCAD dataset\cite{replicad}, randomly cutting it to obtain different environments.

For each type of scenario, we randomly sample one million environment–trajectory pairs as the training set and evaluate the model in one thousand unseen environments. All training and evaluations are run on Ubuntu 20.04 with an Intel i9-14900 CPU and a GeForce RTX 4090D GPU. We use the Adam\cite{adam} optimizer with a learning rate 1.0e-4 to update the model parameters. The weights of $\mathcal{L}_{safe}$, $\mathcal{L}_{smooth}$, $\mathcal{L}_{unip}$ are set to $50.0$, $0.1$, and $20.0$, respectively. We stack $2$ cascade diffusion transformer layers\cite{dit} for PTDM, where the number of the primitives is set to $32$, the sequence length of the robot path $N_{\boldsymbol{\tau}}=64$. Following the configuration for anchor-based diffusion\cite{diffusion_drive}, the training diffusion schedule is truncated by $50/1200$ to diffuse the primitive. During inference, only $2$ denoising steps are used for evaluation.

\subsection{Evaluation Metrics}
We use following four kinds of metrics to evaluate the performance of the planner.
\begin{itemize}
\item \textit{Diversity score of the paths} (D.S.):
Following the work~\cite{diffusion_drive}, we use the method based on mean Intersection over Union (mIoU) to evaluate the diversity of the paths generated by the model. Specifically, the diversity score of a single sampling result is
\begin{align}
D.S. = 1-\frac{1}{N}\sum_{i=1}^{N}\frac{\text{Vol}(\mathfrak{T}_i\cap\bigcup_{j=1}^N\mathfrak{T}_j)}{\text{Vol}(\mathfrak{T}_i\cup\bigcup_{j=1}^N\mathfrak{T}_j)},
\end{align}
where $\mathfrak{T}_i,i=1,2,...,N$ represents the swept body of $i$-th denoised trajectory, $\bigcup_{j=1}^N\mathfrak{T}_j$ is the union of all of them. The function $\text{Vol}(\cdot)$ is used to calculate volume. A higher score indicates higher diversity. In this work, to evaluate more efficiently, we approximate the swept body by the collision model of the DDMoMa.
\item \textit{Success rate of the planning} (S.R.$/\%$): 
Planning attempts will be considered failed if either the planner cannot find feasible path within limited time or none of the optimization processes successfully return with dynamically feasible and collision-free trajectories.
\item \textit{Time consumption of planning}  (T.P.$/ms$): 
It consists of the time consumed by the path finding in frontend and optimization-based backend.
\item \textit{Trajectory duration} (T.D.$/s$): 
The total time required for a robot to execute a planned motion from start to goal, which reflects the optimality of the trajectory.
\end{itemize}

\subsection{Quantitative Evaluation}
In the following experiments, we seek to evaluate the following claims: 
\begin{itemize}
\item \textit{Claim 1.}
Proposed primitive-based truncated diffusion can enhance the efficiency and diversity of the paths.
\item \textit{Claim 2.}
Proposed task representation encoder that combines the attention mechanisms with KSE can achieve more effective feature fusion for the environment point clouds and the boundary states of DDMoMa, thereby improving the success rate and optimality of planning.
\item \textit{Claim 3.}
Proposed learning-enhanced planner can achieve higher success rate and efficiency than state-of-the-art (SoTA) classical method.
\end{itemize}

To validate \textit{Claim 1}, we conduct comparisons of diversity and time consumption between PTDM,  vanilla\cite{ddpm} and anchor-based\cite{diffusion_drive} diffusion models. Fig.~\ref{fig:line_ptdm} shows the line charts of average T.P and D.S. regarding the sample size of the diffusion model. As shown in Fig.~\ref{fig:line_ptdm}(a), PTDM significantly enhances the diversity of the sampled results. We attribute this improvement to the clearer distribution provided to the denoising module, which more effectively prevents mode collapse\cite{diffusion_drive}. As shown in Fig.~\ref{fig:line_ptdm}(b), by leveraging DDIM, which greatly reduces the number of inference steps, PTDM achieves an order-of-magnitude improvement in efficiency compared with vanilla diffusion model. The T.P. of anchor-based diffusion is longer than that of PTDM and increases sharply as the sample size grows from $4$ to $5$. This is mainly because it requires sampling for all primitives, resulting in substantial computational cost. Interestingly, the T.P. of PTDM slightly decreases as the number of samples increases. This phenomenon also occurs in anchor-based diffusion in \textit{Mixed} and \textit{Replica}, when the sample size increases from $3$ to $4$. We speculate that the greater variety of trajectories provides a higher probability of good initialization, thereby reducing optimization time consumption and enabling earlier termination of post-processing.

\begin{figure}[t]  
		\centering
		{\includegraphics[width=1.0\columnwidth]{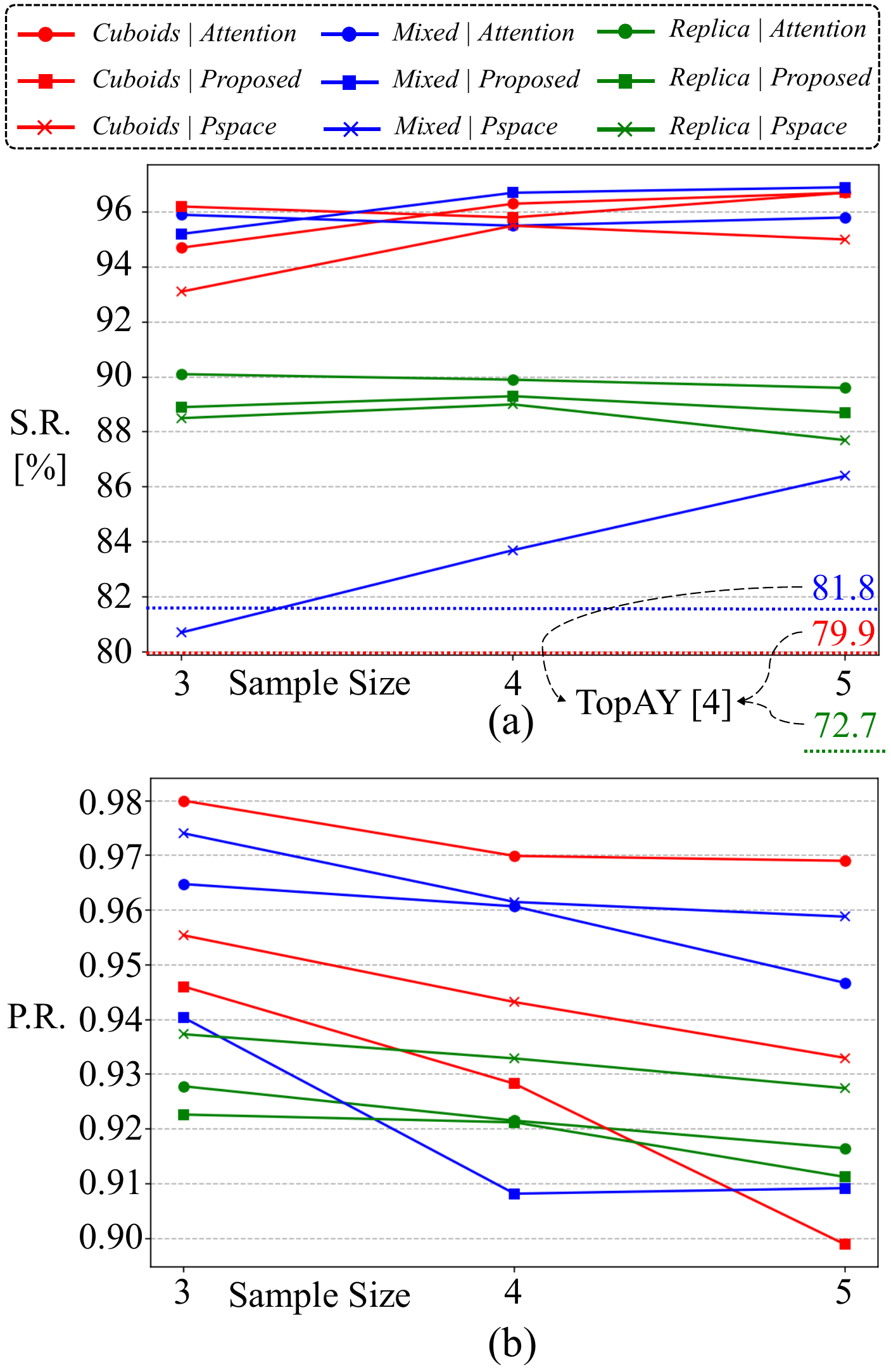}}
		\caption{Ablation study on task encoders: success rate (S.R.) and path ratio (P.R.) in three types of environments. The \textcolor[rgb]{1.0,0.0,0.0}{red}, \textcolor[rgb]{0.0,0.0,1.0}{blue}, and \textcolor[rgb]{0.0,0.5,0.0}{green} dashed lines in subfigure (a) represent the average S.R. of TopAY\cite{topay} in \textit{Cuboids}, \textit{Mixed}, and \textit{Replica}, respectively.}
		\label{fig:line_fuse}
\end{figure}

To validate \textit{Claim 2}, we conduct comparisons between proposed task encoder and the following task encoders:
\begin{itemize}
\item \textit{Pspace}: 
KSE is employed, but instead of encoding the keypoint sequence separately, it is integrated into the obstacle point cloud. A new four-dimensional feature $[o,e_1,e_2,e_3]$ is added to distinguish keypoints from obstacles. Here, $o=0,1,...,N_{m}+1$ represents the keypoint order, while obstacle points are assigned a value of $-100$ in this dimension. Using one-hot encoding, $[e_1,e_2,e_3]=[1,0,0]\vee[0,1,0]\vee[0,0,1]$ denote the obstacles, start state, and goal state, respectively.
\item \textit{Attention}: 
KSE is not used. Instead, MLP is adopted directly for feature extraction of boundary states. These features are then fused with the features from obstacle point cloud through the cross attention mechanism.
\end{itemize}
The results are shown in Fig.~\ref{fig:line_fuse}, where the path ratio (P.R.) denotes the ratio between the T.D. of the learning-enhanced planner and that of SoTA classical method, TopAY\cite{topay}, in the same task. A lower P.R. indicates better optimality, and a P.R. less than $1$ signifies that the result outperforms TopAY\cite{topay}. As shown in Fig.~\ref{fig:line_fuse}, the proposed task encoder achieves the highest success rate in most cases in \textit{Cuboids} and \textit{Mixed}, and also outperforms other encoders in terms of P.R. 

Although \textit{Pspace} incorporates boundary states into the task encoder and maps them into the obstacle point cloud space, their substantial differences in physical meaning may cause the encoder designed specifically for point clouds to lose useful information during downsampling, making it difficult to extract effective features directly from the fused point cloud. In contrast, the proposed method encodes each keypoint using MLP without downsampling and fuses features with the obstacles through a cross attention mechanism, achieving superior performance. While \textit{Attention} and \textit{Proposed} achieve comparable S.R. across different environments, \textit{Proposed} consistently achieves a considerable advantage over \textit{Attention} in terms of optimality, as indicated by lower P.R., further validating the effectiveness of KSE. 

To validate \textit{Claim 3}, we compare the proposed planner with SoTA classical method, TopAY\cite{topay}. As shown in Fig.~\ref{fig:line_ptdm}(b) and Fig.~\ref{fig:line_fuse}, our method outperforms TopAY\cite{topay} in terms of success rate, efficiency, and optimality, demonstrating the effectiveness and efficiency of the proposed framework in addressing the motion planning problem of DDMoMa. More demonstrations and more detailed comparisons between different task encoders and diffusion models can be found on our project page\footnote{\url{https://nmoma.github.io/nmoma/}}.

\section{Conclusion \& Limitations}
In this paper, we present a learning-enhanced motion planning framework for DDMoMa. Benefiting from proposed KSE and PTDM, in cluttered 3D simulation environments, our method achieves higher success rates and trajectory diversity compared to vanilla diffusion models and classical baselines, while maintaining competitive runtime efficiency.

Despite these promising results, several limitations remain to be addressed for a broader application. Our validation is currently restricted to static simulation environments, and the gap between simulation and the physical world, which is characterized by perception noise and control uncertainty, has not yet been tested. Limited onboard computing resources may also lead to reduced efficiency. Furthermore, our current diffusion process is conducted at discrete path points, still requiring trajectory optimization-based post-processing to obtain dynamically feasible trajectories, thereby increasing the complexity of the framework.

In the future, we will try to incorporate advanced diffusion frameworks, such as EDM\cite{edm}, Consistency model\cite{consistency_model}, and perform diffusion directly in the trajectory representation space to further accelerate inference and improve success rate\cite{gpd}. We also plan to deploy the planner on physical robots to test and bridge the potential sim-to-real gap. To further exploit the advantages of the efficiency of our method, we will also consider adapting it to mobile manipulators with different configurations and more environments with different levels of difficulty.

\newlength{\bibitemsep}\setlength{\bibitemsep}{0.0\baselineskip}
\newlength{\bibparskip}\setlength{\bibparskip}{0pt}
\let\oldthebibliography\thebibliography
\renewcommand\thebibliography[1]{%
	\oldthebibliography{#1}%
	\setlength{\parskip}{\bibitemsep}%
	\setlength{\itemsep}{\bibparskip}%
}
\bibliography{references}

@article{momatase,
  title={Manipulator motion planning for part pickup and transport operations from a moving base},
  author={Thakar, Shantanu and Rajendran, Pradeep and Kabir, Ariyan M and Gupta, Satyandra K},
  journal={IEEE Transactions on Automation Science and Engineering},
  volume={19},
  number={1},
  pages={191--206},
  year={2020},
  publisher={IEEE}
}

@inproceedings{topo,
  title={Robust real-time uav replanning using guided gradient-based optimization and topological paths},
  author={Zhou, Boyu and Gao, Fei and Pan, Jie and Shen, Shaojie},
  booktitle={2020 IEEE international conference on robotics and automation (ICRA)},
  pages={1208--1214},
  year={2020},
  organization={IEEE}
}

@article{dp,
  title={Diffusion policy: Visuomotor policy learning via action diffusion},
  author={Chi, Cheng and Xu, Zhenjia and Feng, Siyuan and Cousineau, Eric and Du, Yilun and Burchfiel, Benjamin and Tedrake, Russ and Song, Shuran},
  journal={The International Journal of Robotics Research},
  volume={44},
  number={10-11},
  pages={1684--1704},
  year={2025},
  publisher={Sage Publications Sage UK: London, England}
}

@inproceedings{ddim,
    title={Denoising Diffusion Implicit Models},
    author={Jiaming Song and Chenlin Meng and Stefano Ermon},
    booktitle={International Conference on Learning Representations},
    year={2021},
    url={https://openreview.net/forum?id=St1giarCHLP}
}

@article{edm,
  title={Elucidating the design space of diffusion-based generative models},
  author={Karras, Tero and Aittala, Miika and Aila, Timo and Laine, Samuli},
  journal={Advances in neural information processing systems},
  volume={35},
  pages={26565--26577},
  year={2022}
}

@inproceedings{consistency_model,
  author={Yang Song and Prafulla Dhariwal and Mark Chen and Ilya Sutskever},
  title={Consistency Models},
  year={2023},
  cdate={1672531200000},
  pages={32211-32252},
  url={https://proceedings.mlr.press/v202/song23a.html},
  booktitle={ICML},
}

@inproceedings{truncated_diffusion,
    title={Truncated Diffusion Probabilistic Models and Diffusion-based Adversarial Auto-Encoders},
    author={Huangjie Zheng and Pengcheng He and Weizhu Chen and Mingyuan Zhou},
    booktitle={The Eleventh International Conference on Learning Representations },
    year={2023},
    url={https://openreview.net/forum?id=HDxgaKk956l}
}

@article{ompl,
  title={The open motion planning library},
  author={Sucan, Ioan A and Moll, Mark and Kavraki, Lydia E},
  journal={IEEE Robotics \& Automation Magazine},
  volume={19},
  number={4},
  pages={72--82},
  year={2012},
  publisher={IEEE}
}

@inproceedings{focal_loss,
  title={Focal loss for dense object detection},
  author={Lin, Tsung-Yi and Goyal, Priya and Girshick, Ross and He, Kaiming and Doll{\'a}r, Piotr},
  booktitle={Proceedings of the IEEE international conference on computer vision},
  pages={2980--2988},
  year={2017}
}

@inproceedings{ichter2018learning,
  title={Learning sampling distributions for robot motion planning},
  author={Ichter, Brian and Harrison, James and Pavone, Marco},
  booktitle={2018 IEEE International Conference on Robotics and Automation (ICRA)},
  pages={7087--7094},
  year={2018},
  organization={IEEE}
}

@article{replica,
  title={The replica dataset: A digital replica of indoor spaces},
  author={Straub, Julian and Whelan, Thomas and Ma, Lingni and Chen, Yufan and Wijmans, Erik and Green, Simon and Engel, Jakob J and Mur-Artal, Raul and Ren, Carl and Verma, Shobhit and others},
  journal={arXiv preprint arXiv:1906.05797},
  year={2019}
}

@inproceedings{dit,
  title={Scalable diffusion models with transformers},
  author={Peebles, William and Xie, Saining},
  booktitle={Proceedings of the IEEE/CVF international conference on computer vision},
  pages={4195--4205},
  year={2023}
}

@article{adam,
  title={Adam: A method for stochastic optimization},
  author={Kingma, Diederik P and Ba, Jimmy},
  journal={arXiv preprint arXiv:1412.6980},
  year={2014}
}

@inproceedings{replicad,
    title     =     {Habitat 2.0: Training Home Assistants to Rearrange their Habitat},
    author    =     {Andrew Szot and Alex Clegg and Eric Undersander and Erik Wijmans and Yili Zhao and John Turner and Noah Maestre and Mustafa Mukadam and Devendra Chaplot and Oleksandr Maksymets and Aaron Gokaslan and Vladimir Vondrus and Sameer Dharur and Franziska Meier and Wojciech Galuba and Angel Chang and Zsolt Kira and Vladlen Koltun and Jitendra Malik and Manolis Savva and Dhruv Batra},
    booktitle   =     {Advances in Neural Information Processing Systems (NeurIPS)},
    year      =     {2021}
}

@article{mpnet,
  title={Motion planning networks: Bridging the gap between learning-based and classical motion planners},
  author={Qureshi, Ahmed Hussain and Miao, Yinglong and Simeonov, Anthony and Yip, Michael C},
  journal={IEEE Transactions on Robotics},
  volume={37},
  number={1},
  pages={48--66},
  year={2020},
  publisher={IEEE}
}

@inproceedings{ptrans3,
  title={Point transformer v3: Simpler faster stronger},
  author={Wu, Xiaoyang and Jiang, Li and Wang, Peng-Shuai and Liu, Zhijian and Liu, Xihui and Qiao, Yu and Ouyang, Wanli and He, Tong and Zhao, Hengshuang},
  booktitle={Proceedings of the IEEE/CVF conference on computer vision and pattern recognition},
  pages={4840--4851},
  year={2024}
}

@ARTICLE{MPD2025,
  author={Carvalho, João and Le, An Thai and Kicki, Piotr and Koert, Dorothea and Peters, Jan},
  journal={IEEE Transactions on Robotics}, 
  title={Motion Planning Diffusion: Learning and Adapting Robot Motion Planning With Diffusion Models}, 
  year={2025},
  volume={41},
  number={},
  pages={4881-4901},
  doi={10.1109/TRO.2025.3593109}}

@article{hzc_sr,
  title={Hierarchically depicting vehicle trajectory with stability in complex environments},
  author={Han, Zhichao and Tian, Mengze and Gongye, Zaitian and Xue, Donglai and Xing, Jiaxi and Wang, Qianhao and Gao, Yuman and Wang, Jingping and Xu, Chao and Gao, Fei},
  journal={Science Robotics},
  volume={10},
  number={103},
  pages={eads4551},
  year={2025},
  publisher={American Association for the Advancement of Science}
}

@article{rampage,
  title={RAMPAGE: Toward whole-body, real-time, and agile motion planning in unknown cluttered environments for mobile manipulators},
  author={Yang, Yuqiang and Meng, Fei and Meng, Zehui and Yang, Chenguang},
  journal={IEEE Transactions on Industrial Electronics},
  volume={71},
  number={11},
  pages={14492--14502},
  year={2024},
  publisher={IEEE}
}

@inproceedings{zzd_icra,
  title={Real-time whole-body motion planning for mobile manipulators using environment-adaptive search and spatial-temporal optimization},
  author={Wu, Chengkai and Wang, Ruilin and Song, Mianzhi and Gao, Fei and Mei, Jie and Zhou, Boyu},
  booktitle={2024 IEEE International Conference on Robotics and Automation (ICRA)},
  pages={1369--1375},
  year={2024},
  organization={IEEE}
}

@article{m2diffuser,
  title={M 2 diffuser: Diffusion-based trajectory optimization for mobile manipulation in 3d scenes},
  author={Yan, Sixu and Zhang, Zeyu and Han, Muzhi and Wang, Zaijin and Xie, Qi and Li, Zhitian and Li, Zhehan and Liu, Hangxin and Wang, Xinggang and Zhu, Song-Chun},
  journal={IEEE Transactions on Pattern Analysis and Machine Intelligence},
  year={2025},
  publisher={IEEE}
}

@article{learn_neural,
  title={Neural randomized planning for whole body robot motion},
  author={Lu, Yunfan and Ma, Yuchen and Hsu, David and Cai, Panpan},
  journal={arXiv preprint arXiv:2405.11317},
  year={2024}
}

@inproceedings{presto,
  title={Presto: Fast motion planning using diffusion models based on key-configuration environment representation},
  author={Seo, Mingyo and Cho, Yoonyoung and Sung, Yoonchang and Stone, Peter and Zhu, Yuke and Kim, Beomjoon},
  booktitle={2025 IEEE International Conference on Robotics and Automation (ICRA)},
  pages={10861--10867},
  year={2025},
  organization={IEEE}
}

@article{ddpm,
  title={Denoising diffusion probabilistic models},
  author={Ho, Jonathan and Jain, Ajay and Abbeel, Pieter},
  journal={Advances in neural information processing systems},
  volume={33},
  pages={6840--6851},
  year={2020}
}

@inproceedings{diffusion_drive,
  title={Diffusiondrive: Truncated diffusion model for end-to-end autonomous driving},
  author={Liao, Bencheng and Chen, Shaoyu and Yin, Haoran and Jiang, Bo and Wang, Cheng and Yan, Sixu and Zhang, Xinbang and Li, Xiangyu and Zhang, Ying and Zhang, Qian and others},
  booktitle={Proceedings of the Computer Vision and Pattern Recognition Conference},
  pages={12037--12047},
  year={2025}
}

@inproceedings{rotation6d,
  title={On the continuity of rotation representations in neural networks},
  author={Zhou, Yi and Barnes, Connelly and Lu, Jingwan and Yang, Jimei and Li, Hao},
  booktitle={Proceedings of the IEEE/CVF conference on computer vision and pattern recognition},
  pages={5745--5753},
  year={2019}
}

@article{mode_collapse,
  title={Avoiding mode collapse in diffusion models fine-tuned with reinforcement learning},
  author={Barcel{\'o}, Roberto and Alc{\'a}zar, Crist{\'o}bal and Tobar, Felipe},
  journal={arXiv preprint arXiv:2410.08315},
  year={2024}
}

@inproceedings{priordm,
  title={Prior does matter: Visual navigation via denoising diffusion bridge models},
  author={Ren, Hao and Zeng, Yiming and Bi, Zetong and Wan, Zhaoliang and Huang, Junlong and Cheng, Hui},
  booktitle={Proceedings of the IEEE/CVF Conference on Computer Vision and Pattern Recognition},
  pages={12100--12110},
  year={2025}
}

@inproceedings{cdm,
  title={Cascaded Diffusion Models for Neural Motion Planning},
  author={Sharma, Mohit and Fishman, Adam and Kumar, Vikash and Paxton, Chris and Kroemer, Oliver},
  booktitle={2025 IEEE International Conference on Robotics and Automation (ICRA)},
  pages={14361--14368},
  year={2025},
  organization={IEEE}
}

@inproceedings{gpd,
  title={GPD: Guided Polynomial Diffusion for Motion Planning},
  author={Srikanth, Ajit and Mahajan, Parth and Saha, Kallol and Mandadi, Vishal and Paul, Pranjal and Wadhwani, Pawan and Bhowmick, Brojeshwar and Singh, Arun and Krishna, Madhava},
  booktitle={2025 IEEE 21st International Conference on Automation Science and Engineering (CASE)},
  pages={2758--2765},
  year={2025},
  organization={IEEE}
}

@inproceedings{
    robocasa,
    title={RoboCasa365: A Large-Scale Simulation Framework for Training and Benchmarking Generalist Robots},
    author={Soroush Nasiriany and Sepehr Nasiriany and Abhiram Maddukuri and Yuke Zhu},
    booktitle={The Fourteenth International Conference on Learning Representations},
    year={2026},
    url={https://openreview.net/forum?id=tQJYKwc3n4}
}

@ARTICLE{e2e_fly,
  author={Han, Zhichao and Xu, Long and Pei, Liuao and Gao, Fei},
  journal={IEEE Robotics and Automation Letters}, 
  title={Dynamically Feasible Trajectory Generation With Optimization-Embedded Networks for Autonomous Flight}, 
  year={2025},
  volume={10},
  number={10},
  pages={9995-10002},
  doi={10.1109/LRA.2025.3595077}}

@article{topay,
  title={TopAY: Efficient Trajectory Planning for Differential Drive Mobile Manipulators via Topological Paths Search and Arc Length-Yaw Parameterization},
  author={Xu, Long and Wong, Choilam and Zhang, Mengke and Lin, Junxiao and Hou, Jialiang and Gao, Fei},
  journal={arXiv preprint arXiv:2507.02761},
  year={2025}
}

\end{document}